\documentclass[letterpaper, 10 pt, conference]{ieeeconf}  

\IEEEoverridecommandlockouts                              

\usepackage{xcolor}
\usepackage{graphicx}
\usepackage{amssymb}
\usepackage{diagbox}
\usepackage{dblfloatfix}
\usepackage{amsmath}
\usepackage{booktabs}
\usepackage{pifont}
\usepackage{subcaption}
\usepackage{comment}
\usepackage{ulem}
\usepackage{cuted} 
\usepackage{caption} 
\usepackage[font=footnotesize,labelsep=period]{caption} 
\usepackage{hyperref}

\newcommand{\cmark}{\checkmark} 
\newcommand{\xmark}{\ding{55}}  

\title{\LARGE \bf
STONE Dataset: A Scalable Multi-Modal Surround-View 3D Traversability Dataset for Off-Road Robot Navigation
}

\author{Konyul Park$^{1}$$^{*}$, Daehun Kim$^{2}$$^{*}$, Jiyong Oh$^{2}$, Seunghoon Yu$^{2}$, Junseo Park$^{1}$, Jaehyun Park$^{2}$, Hongjae Shin$^{1}$, \\ Hyungchan Cho$^{1}$, Jungho Kim$^{1}$, and Jun Won Choi$^{2}$$^{\dagger}$
\thanks{$^{*}$These authors contributed equally to this work.}
\thanks{$^{\dagger}$Corresponding author.}
\thanks{$^{1}$Interdisciplinary Program in Artificial Intelligence, Seoul National University,
Seoul, 08826, Korea.}
\thanks{$^{2}$Department of Electrical and Computer Engineering, Seoul National University,
Seoul, 08826, Korea.}
\thanks{{\tt\footnotesize \{kypark, dhkim, jyoh, shyu, jspark, jhpark, hjshin, hccho, jhkim\}@adr.snu.ac.kr}}
\thanks{{\tt\footnotesize junwchoi@snu.ac.kr}}
}

\begin{document}

\maketitle

\thispagestyle{empty}
\pagestyle{empty}

\begin{strip}
\vspace*{-30mm}   
\centering
\includegraphics[width=\textwidth]{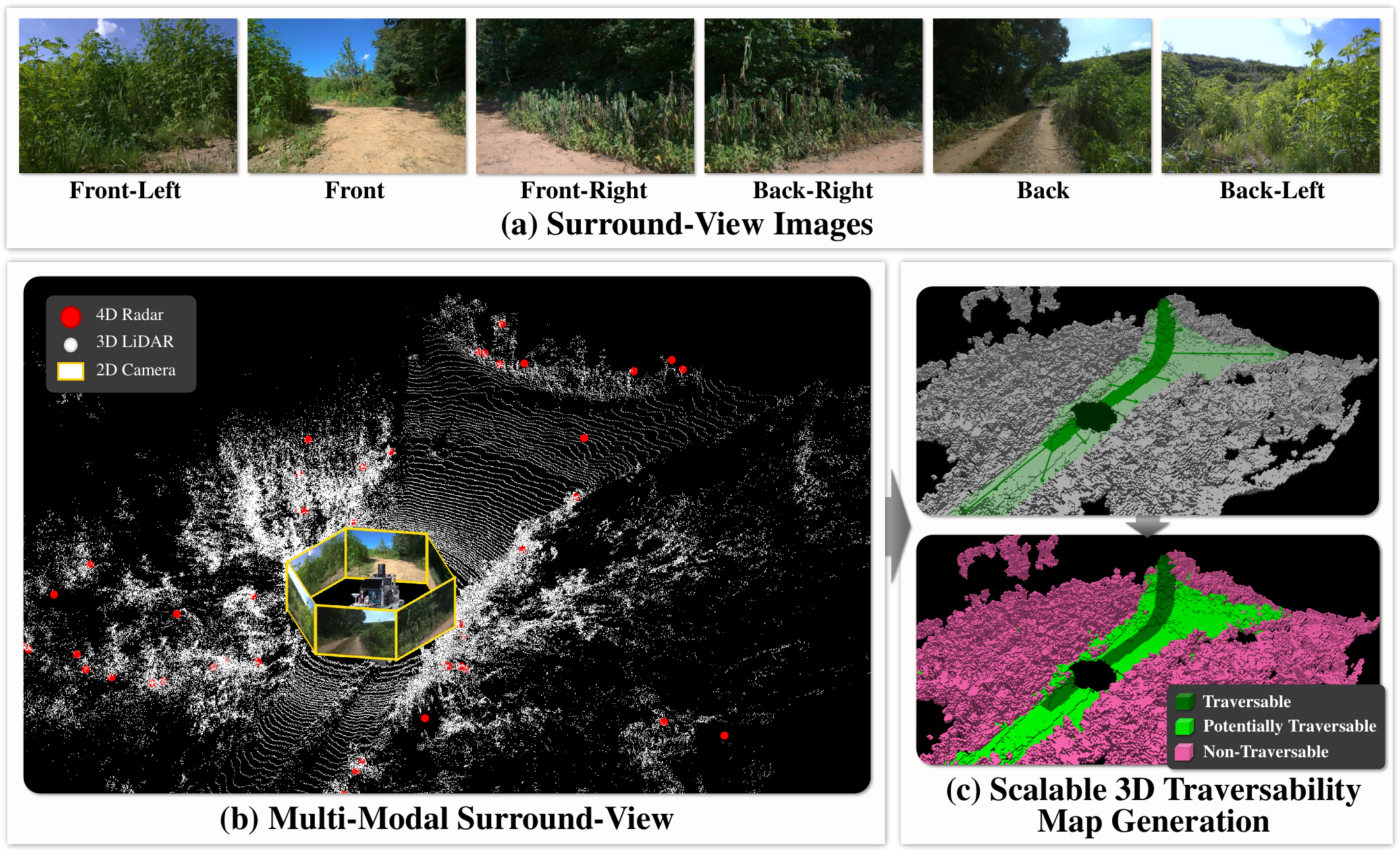}
\captionof{figure}{\textbf{Overview of the STONE dataset.} The STONE dataset is a multi-modal 3D traversability dataset collected in off-road environments, which provides ground-truth annotations of 3D traversable areas automatically without human effort. 
(a) illustrates surround-view images captured around the robot.
(b) shows the 3D scene captured by the multi-modal surround-view obtained using LiDAR, cameras, and 4D radar. 
(c) presents the process of automatically generating scalable 3D traversability maps based on robot trajectories.
}
\label{fig:intro}
\end{strip}

\begin{abstract}

Reliable off-road navigation requires accurate estimation of traversable regions and robust perception under diverse terrain and sensing conditions. However, existing datasets lack both scalability and multi-modality, which limits progress in 3D traversability prediction. In this work, we introduce STONE, a large-scale multi-modal dataset for off-road navigation. STONE provides (1) trajectory-guided 3D traversability maps generated by a fully automated, annotation-free pipeline, and (2) comprehensive surround-view sensing with synchronized 128-channel LiDAR, six RGB cameras, and three 4D imaging radars. The dataset covers a wide range of environments and conditions, including day and night, grasslands, farmlands, construction sites, and lakes. Our auto-labeling pipeline reconstructs dense terrain surfaces from LiDAR scans, extracts geometric attributes such as slope, elevation, and roughness, and assigns traversability labels beyond the robot’s trajectory using a Mahalanobis-distance-based criterion. This design enables scalable, geometry-aware ground-truth construction without manual annotation. Finally, we establish a benchmark for voxel-level 3D traversability prediction and provide strong baselines under both single-modal and multi-modal settings. STONE is available at:
\href{https://konyul.github.io/STONE-dataset/}{https://konyul.github.io/STONE-dataset/}

\end{abstract}

\section{INTRODUCTION}

\begin{table*}[!t]
\centering
\caption{Comparison of several off-road datasets}
\label{tab:offroad_comparison}
\renewcommand{\arraystretch}{1.1}
\setlength{\tabcolsep}{2pt}
\resizebox{\textwidth}{!}{%
\begin{tabular}{lccccccc}
\toprule[1.5pt]
\textbf{Dataset} & \textbf{Sensors} & \textbf{4D Radar} & \textbf{Camera FOV} & \textbf{\# of Cameras} & \textbf{Image Resolution} & \textbf{LiDAR Resolution} & \textbf{Annotation} \\
\midrule
Freiburg Forest \cite{Freiburg} & Camera, NIR & \xmark & Front-view & 2 & 1024$\times$768 & - & 2D semantic \\
YCOR \cite{YCOR}              & Camera, LiDAR, INS & \xmark & Front-view & 1 & 1024$\times$544 & 64 & 2D semantic \\
RUGD \cite{RUGD}               & Camera, LiDAR, IMU, GPS & \xmark & Front-view & 1 & 688$\times$550 & 32 & 2D semantic \\
RELLIS-3D \cite{rellis3d}    & Camera, LiDAR, INS & \xmark & Front-view & 1 & 1920$\times$1200 & 32, 64 & 2D/3D semantic \\
ORFD \cite{orfd}                  & Camera, LiDAR & \xmark & Front-view & 1 & 1280$\times$720 & 40 & 2D traversability \\
TartanDrive 2.0 \cite{tartanv2}   & Camera, LiDAR, INS & \xmark & Front-view & 1 & 1024$\times$512 & 2$\times$32, 70 & - \\
GOOSE \cite{GOOSE}             & Camera, NIR, LiDAR, INS & \xmark & Front-view* & 4 & 2048$\times$1000 &  2$\times$32, 128 & 2D/3D semantic \\
TOMD \cite{tomd}                  & Camera, LiDAR, INS & \xmark & Front-view & 1 & 1920$\times$1080 & 128 & 2D traversability \\
\midrule
\textbf{STONE (Ours)} & \textbf{4D Radar, Camera, LiDAR, INS} & \cmark & \textbf{Surround-view} & \textbf{6} & \textbf{1920$\times$1200} & \textbf{128} & \textbf{3D traversability} \\
\bottomrule[1.5pt]
\multicolumn{8}{l}{*indicates papers that employ a multi-camera setup but release only front-view images.}
\end{tabular}
}
\end{table*}

Robot navigation in off-road environments plays a vital role in applications ranging from military operations to agriculture, construction, and logistics, enabling safer and more efficient human activities. Reliable navigation in such settings hinges on accurate estimation of traversable regions. Unlike on-road environments, where structural cues such as lane markings, curbs, and road boundaries clearly delineate drivable areas, off-road terrains are unstructured and often deformable (e.g., soil, vegetation, and gravel). These terrains lack well-defined boundaries and provide weaker geometric cues, making the reliable identification of traversable regions more challenging.

Existing off-road datasets remain insufficient for reliable traversability estimation. While RELLIS-3D~\cite{rellis3d} and RUGD~\cite{rugd} provide semantic annotations, such labels are costly to obtain, and semantics alone are insufficient to determine drivability, as regions sharing the same label may differ significantly due to geometric variations. TartanDrive~\cite{tartan} offers top-down height maps, but elevation alone provides limited cues for accurate traversability assessment. ORFD~\cite{orfd} and TOMD~\cite{tomd} rely on manually annotated 2D traversability maps, which are expensive to produce and difficult to scale. Trajectory-based approaches~\cite{ftfoot, vstrong, ssonly} generate self-supervised 2D pixel-level labels from robot paths, while Scate~\cite{scate} derives ground truth solely from vehicle–terrain interaction signals projected onto point clouds, without explicitly exploiting geometric features.
Overall, effective traversability assessment requires rich 3D geometric cues as well as compact representations that can be readily integrated into downstream tasks in the autonomous driving pipeline.

In addition to 3D traversability estimation, reliable off-road navigation requires full-scene 3D perception with comprehensive coverage and resilience under adverse sensing conditions. Maneuvers such as detours, turning, and reversing demand awareness of the entire 360° environment, making forward-view-only sensing insufficient. Radar, in particular, provides robustness under adverse weather where cameras and LiDAR often fail, offering a critical complement for reliable perception. Yet existing datasets~\cite{rellis3d, rugd, YCOR, orfd, tomd} remain limited to front-view cameras or vision–LiDAR modalities, resulting in blind spots and degraded performance under challenging conditions. A detailed comparison is provided in Table~\ref{tab:offroad_comparison}.

To address the limitations of existing datasets and advance research in off-road navigation, we present STONE, a large-scale dataset for off-road robot navigation collected across diverse environments. As shown in Fig.~\ref{fig:intro}, STONE offers (1) 3D traversability maps automatically derived through a scalable, annotation-free pipeline, and (2) comprehensive multi-modal surround-view data that integrates cameras, LiDAR, and 4D radar.

In STONE, we collect data by manually driving an unmanned ground vehicle (UGV) equipped with a high-resolution 128-channel LiDAR, six RGB cameras, three 4D imaging radars, an RTK-capable GNSS, and a high-rate IMU. Each sample is provided with intrinsic and extrinsic calibration parameters. To achieve precise temporal alignment between the cameras and LiDAR, we employ a trigger-based synchronization scheme in which data acquisition is initiated by a LiDAR pulse signal. This design effectively reduces both temporal and spatial misalignment caused by timing offsets between  sensors. As illustrated in Fig.~\ref{fig:offroad_conditions}, the STONE dataset spans a broad spectrum of challenging terrains and includes both daytime and nighttime conditions.

Manual annotation of fine-grained semantic classes is inherently unscalable, requiring substantial human effort and cost. To address this limitation, the STONE dataset provides an automated framework for generating 3D traversability maps. The core idea is to model the geometric attributes of traversable regions based on the areas actually traversed by the robot during data collection. To this end, we represent these geometric attributes using a multivariate Gaussian distribution and estimate the probability of traversability.
The proposed auto-labeling pipeline consists of three stages. First, LiDAR scans are temporally aggregated to reconstruct a dense and wide-area terrain surface. Second, from this surface, we extract geometric cues such as slope, elevation, and roughness, which directly affect vehicle mobility. Third, voxels along the robot’s trajectory are labeled as traversable. Using their feature distribution as a reference, we propagate labels to neighboring voxels by computing the Mahalanobis distance~\cite{mahala}.
This automated pipeline enables scalable construction of traversability maps across diverse environments while maintaining consistent annotation quality.
\begin{figure*}[!t]
    \centering
    \includegraphics[width=\textwidth]{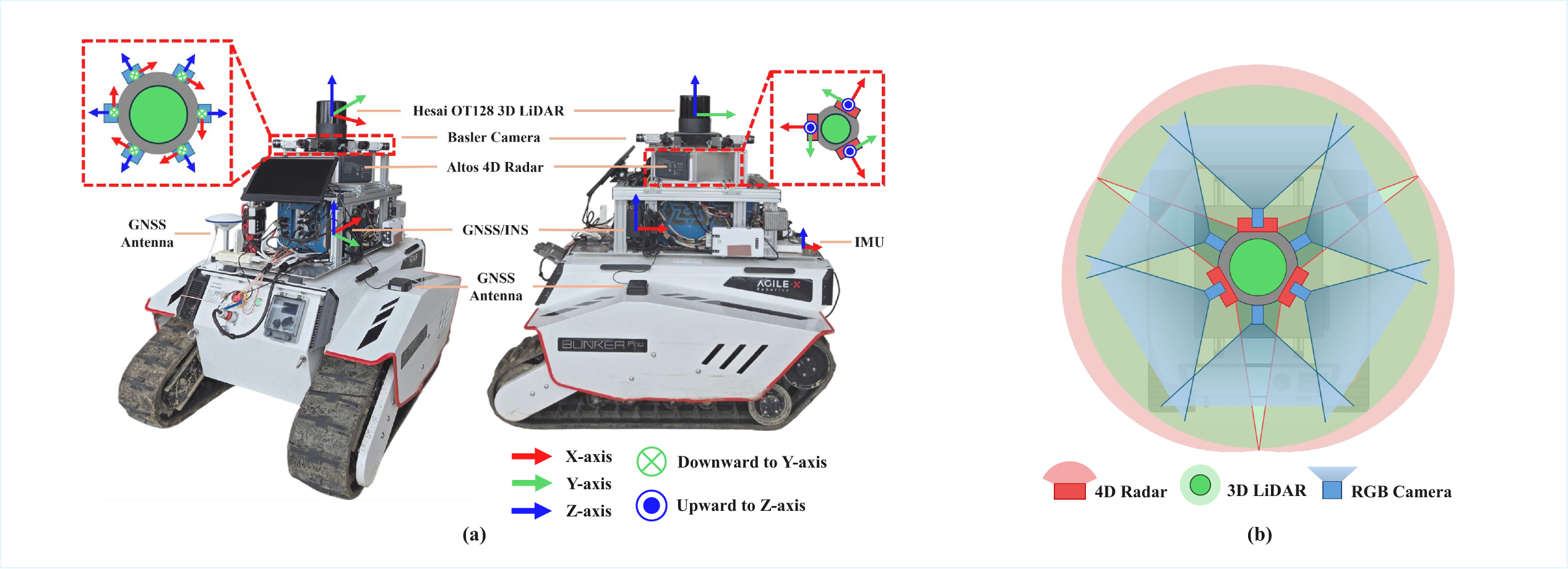}
    \caption{\textbf{Sensor setup and coverage of the Bunker Pro UGV platform.} 
    (a) shows the sensor placement including LiDAR, cameras, 4D radars, IMU and GPS. 
    (b) illustrates the sensing range and coverage in off-road environments.}
    \label{fig:robot_sensor_setup}
\end{figure*}

We establish a benchmark for the 3D traversability prediction task, where the goal is to estimate voxel-level traversability maps from single- or multi-modal sensor inputs. We provide the performance of reference methods for comparative evaluation across three settings: camera-only, LiDAR-only, and multi-modal.

The main contributions of our dataset are summarized as follows:
\begin{itemize}
\item \textbf{Trajectory-guided 3D traversability maps.}
We introduce the first large-scale off-road dataset that provides 3D traversability maps as ground-truth labels. These labels are generated by a fully automated pipeline that reconstructs terrain geometry, extracts mobility-related features (e.g., slope, elevation, and roughness), and propagates traversability information beyond the robot’s trajectory. This approach enables scalable construction of datasets for off-road traversability prediction tasks.

\item \textbf{Comprehensive multi-modal surround-view sensing.}
STONE is the first off-road dataset to integrate synchronized LiDAR, six RGB cameras, and three 4D imaging radars in a surround-view configuration. This setup supports full 360° scene perception and enhances robustness under adverse conditions.

\item \textbf{Diverse environments and conditions.}
The dataset spans diverse terrains (e.g., grasslands, farmlands, construction sites, and lakes) and covers both daytime and nighttime conditions, capturing the real-world variability of off-road environments.

\item \textbf{Benchmark with strong baselines.}
We establish a benchmark for 3D traversability prediction encompassing both single- and multi-modal settings, and provide strong baseline models. This enables standardized evaluation and fosters comparative studies within the community.
\item We will release the dataset publicly.
\end{itemize}

\section{RELATED WORK}

\subsection{Traversability in Off-Road Environments}
 RUGD \cite{rugd}, YCOR \cite{YCOR}, RELLIS-3D \cite{rellis3d}, GOOSE \cite{GOOSE}, and WildOcc \cite{wildocc} have supported progress by providing pixel-, point-, or voxel-level semantic labels for off-road environments. Despite this progress, category-level semantics is insufficient for traversability estimation. For example, regions with the same label (e.g., grass, soil, rubble, or reeds) can vary in traversability depending on elevation, slope, and surface roughness.

ORFD \cite{orfd} and TOMD \cite{tomd} provide manually annotated pixel-level traversability labels. To reduce annotation effort, FtFoot \cite{ftfoot} projects robot trajectories into images to supervise traversability, and V-STRONG \cite{vstrong} combines projected trajectories with SAM-based \cite{SAM} instance masks for pixel-level contrastive costmap learning. Nonetheless, these methods remain limited to 2D supervision. TartanDrive \cite{tartan, tartanv2} provides height maps with images and vehicle dynamics (throttle, steering, IMU), but these signals are insufficient to serve as ground truth for traversability.


\subsection{Sensor Coverage in Off-Road Datasets}
Representative on-road datasets \cite{nuscenes, waymo, argoverse, sit, mantruck} offer multi-modal sensing through surround-view cameras, multiple radars, and 360° spinning LiDAR, providing comprehensive scene coverage. In contrast, as summarized in Table \ref{tab:offroad_comparison}, most off-road datasets \cite{Freiburg, YCOR, RUGD, rellis3d, orfd, tartanv2, GOOSE, tomd} are limited to a front-view camera and at most a single LiDAR, with no incorporation of radar sensing that is crucial under adverse weather.

\begin{figure*}[!t]
    \centering
    \includegraphics[width=0.9\textwidth]{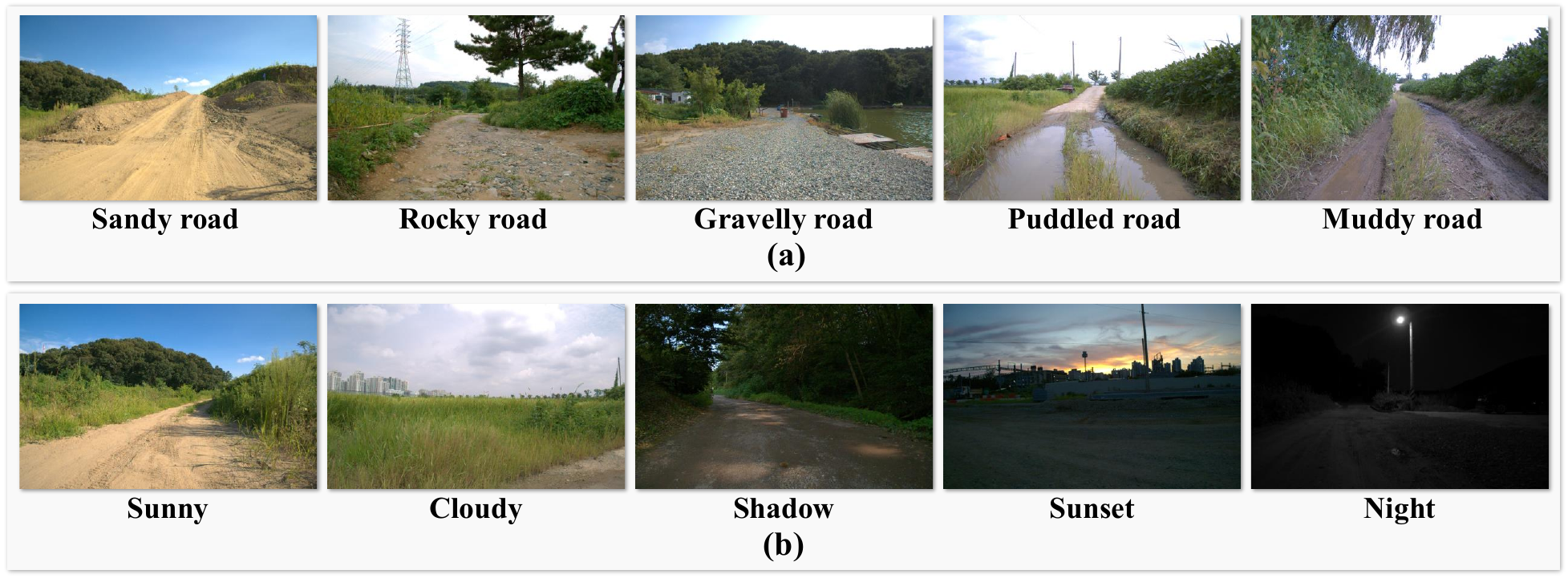}
    \caption{\textbf{Various conditions in which the dataset was collected.} 
    (a) illustrates off-road terrains such as sandy, rocky, gravelly, puddled, and muddy. 
    (b) shows diverse illumination conditions in the dataset: sunny, cloudy, shadow, sunset, and night.}
    \label{fig:offroad_conditions}
\end{figure*}

\section{STONE DATASET}

\subsection{Sensor Setup}
\label{section:setup}
We collected data across diverse off-road environments using the Bunker Pro platform~\cite{bunkerpro}, a tracked UGV  designed for versatile industrial applications. As illustrated in Fig.~\ref{fig:robot_sensor_setup}, the platform was equipped with multiple sensors, providing 360° perception coverage without blind spots.
\begin{itemize}
  \item \textbf{360$^{\circ}$ Rotating LiDAR}: 1 $\times$ Hesai OT128 with 128 channels, a maximum range of 200 m, a field of view of 360$^{\circ}$ (H) $\times$ 40$^{\circ}$ (V), an angular resolution of 0.1$^{\circ}$ (H) $\times$ 0.125$^{\circ}$ (V), and a scanning frequency of 10 Hz.

\item \textbf{Multi-view RGB Cameras}: 6 $\times$ Basler ACE2 2A1920-51gcPRO with a resolution of 1920 $\times$ 1200 and a frame rate of 10 Hz.

\item \textbf{4D Imaging Radars}: 3 $\times$ Continental ARS 548 RDI with a scanning frequency of 20 Hz.

\item \textbf{Global Navigation Satellite System (GNSS)}: NovAtel PIM222A dual-antenna GNSS/INS with RTK capability and an update rate of 20 Hz.

\item \textbf{Inertial Measurement Unit (IMU)}: EPSON G366P IMU providing inertial measurements at 200 Hz.
\end{itemize}
The multi-modal system was integrated and operated on Ubuntu 22.04 using the ROS 2 Humble framework.

\subsection{Sensor Calibration}
\label{section:calib}
The STONE dataset provides both extrinsic and intrinsic calibration parameters. To ensure consistency across coordinate frames, all extrinsic parameters are defined with respect to the LiDAR reference frame. Camera–LiDAR and radar–LiDAR calibrations were performed using open-source tools \cite{velo2cam, radar_calib}, while the IMU–LiDAR extrinsic translation was obtained by measuring the relative sensor positions, and the relative rotation was set to the identity, assuming a co-aligned mounting. We calibrated the intrinsic parameters of the six cameras using a checkerboard-based method \cite{opencalib}.

\subsection{Time Synchronization}
\label{section:timesync}
As the LiDAR spins, each camera is triggered when the LiDAR scan reaches the azimuth angle corresponding to the camera’s viewing direction. To this end, we employed a custom trigger board that converted LiDAR pulse signals into angle-specific triggers to control camera shutters, aligning image captures with LiDAR scan angles and maintaining temporal synchronization. All other sensors were synchronized to LiDAR timestamps: RTK signals were linearly interpolated in $(x,y,z)$ coordinates as well as  quaternions \cite{slerp}, while radar and IMU measurements were temporally aligned by selecting the samples closest to each LiDAR frame stamp.


\begin{figure}[t]
    \centering
    \includegraphics[width=0.8\linewidth]{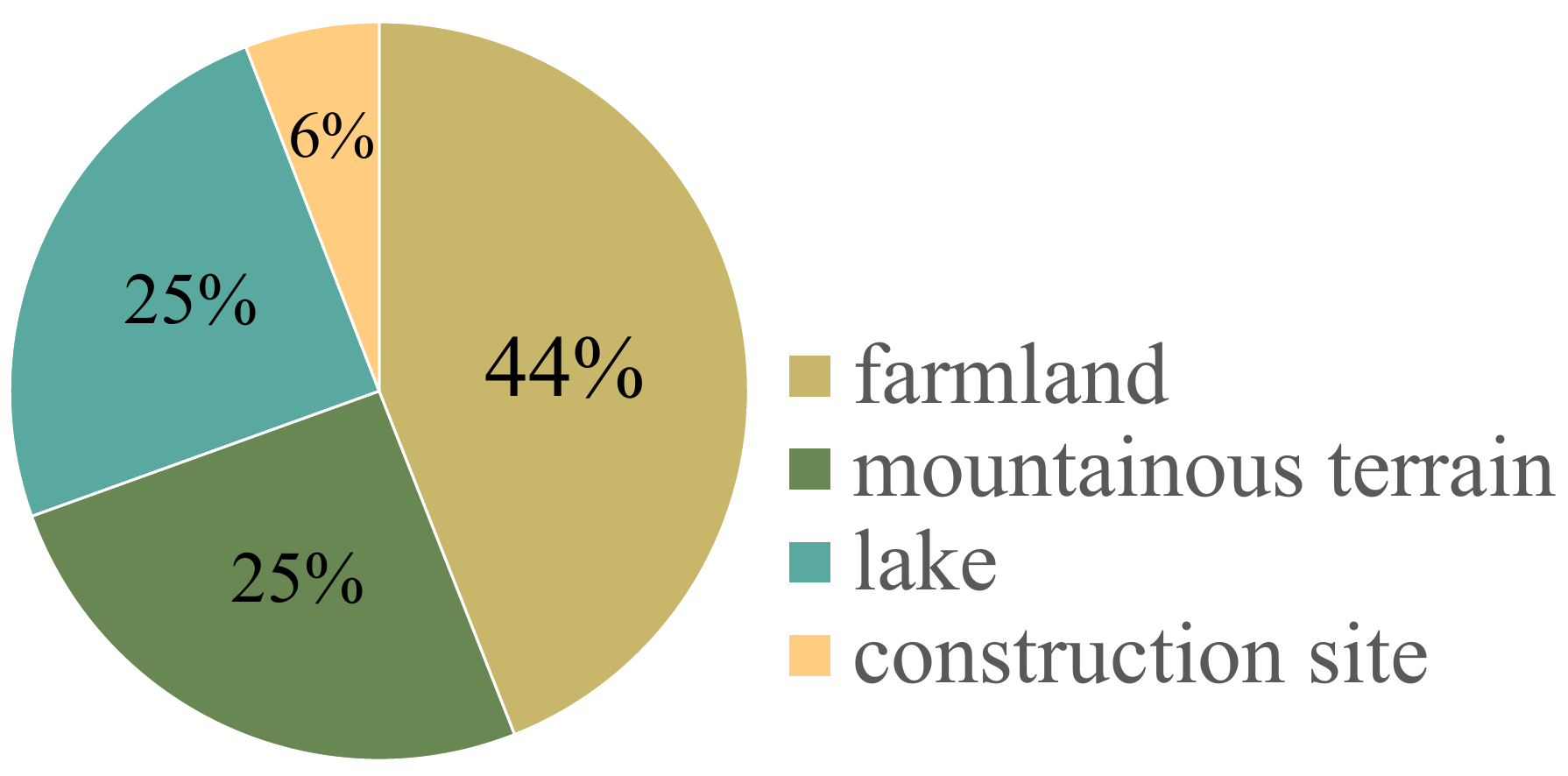}
    \caption{Dataset composition across different environments.}
    \label{fig:dataset_graph}
\end{figure}

\subsection{Data Collection Environments}
\label{section:collect}
The STONE dataset was collected in rural areas surrounding Seoul, South Korea. As illustrated in Fig.~\ref{fig:offroad_conditions}, it encompasses diverse off-road scenarios with varying levels of difficulty, including irregular unpaved roads along lakesides and narrow paddy embankments. The dataset also includes construction sites with heavy machinery and materials, as well as densely vegetated areas where drivable boundaries are poorly defined. In addition to daytime runs, we collected nighttime sequences to evaluate algorithm robustness under varying illumination conditions.



\begin{figure*}[!t]
    \centering
    \includegraphics[width=\textwidth]{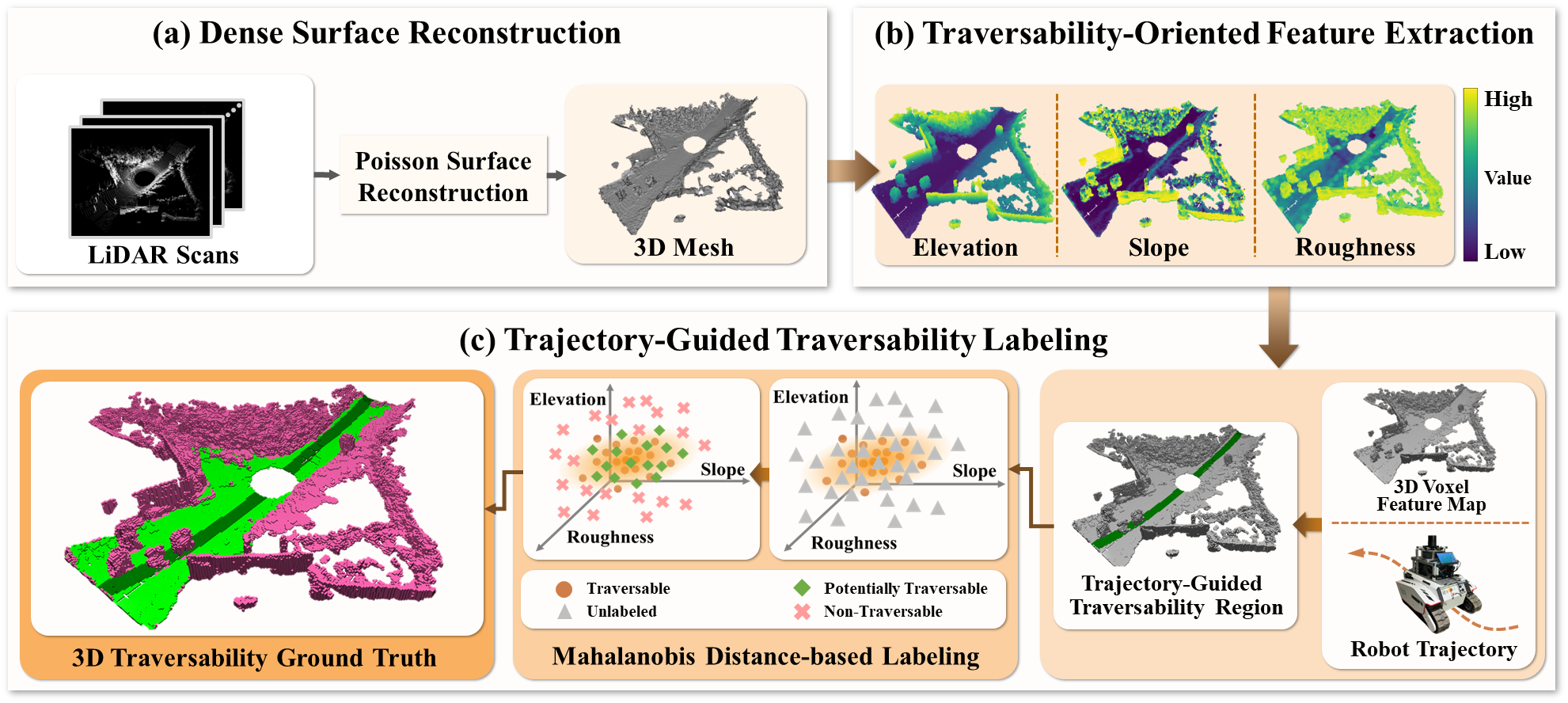}
    \caption{\textbf{Overview of automated 3D traversability map generation process.} (a) LiDAR point clouds are accumulated over multiple time steps in the global coordinate system and reconstructed into a 3D mesh using Poisson surface reconstruction. (b) For each vertex of the 3D mesh, geometric features such as elevation, slope, and roughness are extracted to construct a traversability map. (c) A reference distribution is computed from the geometric features of voxels along the robot’s driving trajectories using a multivariate Gaussian, and a 3D traversability ground-truth label is automatically generated for each voxel whose Mahalanobis distance falls below a predefined threshold.}
    \label{fig:gt_gen}
\end{figure*}

\subsection{Dataset Composition and Trainval Split}
\label{section:statistic}
The dataset consists of 43 sequences with a total of 50,878 frames. 
Each sequence denotes a temporally continuous run of the robot recorded with the full multi-modal sensor suite at 10 Hz in a specific environment or route. 
Sequences are at least 20 seconds long to ensure sufficient temporal context for learning and evaluation. 
The data was collected across four environments: 
farmland, mountainous terrain, lakes, and construction sites. The collected sequences are distributed across these environments, as illustrated in Fig.~\ref{fig:dataset_graph}.
For benchmarking, the dataset is allocated into 31 sequences ($\approx$36,000 frames) for training, 7 sequences ($\approx$8,000 frames) for validation, and 5 sequences ($\approx$7,000 frames) for testing. The split is structured to ensure representation of both daytime and nighttime conditions as well as diverse terrain types in each set. The test set was collected in regions different from those used for the training set.


\section{3D Traversability Map Generation}
\label{section:traverGT}
In this section, we present a scalable framework that automatically generates 3D traversability map from LiDAR data and robot trajectory records. 
As illustrated in Fig.~\ref{fig:gt_gen}, the proposed GT generation pipeline consists of three stages: (i) dense surface reconstruction, (ii) traversability-oriented feature extraction, and (iii) trajectory-guided traversability auto-labeling.

\subsection{Dense Surface Reconstruction}
Sparse and noisy LiDAR returns often hinder the extraction of reliable geometric features. To mitigate this issue, we aggregate multiple consecutive LiDAR scans during surface reconstruction. Specifically, individual LiDAR frames are aligned in a global coordinate frame using the robot’s odometry and fused into a denser point cloud. This aggregation increases point density and yields a more complete representation of the 3D scene.
We then apply the Poisson surface reconstruction method \cite{poisson} to convert the aggregated point cloud into a watertight 3D mesh, which fills in missing regions and suppresses noise. The reconstructed 3D mesh is defined as $M = (\mathcal{V}, \mathcal{F})$,
where $\mathcal{V}$ and $\mathcal{F}$ denote the sets of vertices and faces, respectively.

\begin{figure*}[!t]
    \centering
        \centering
        \includegraphics[width=0.93\textwidth]{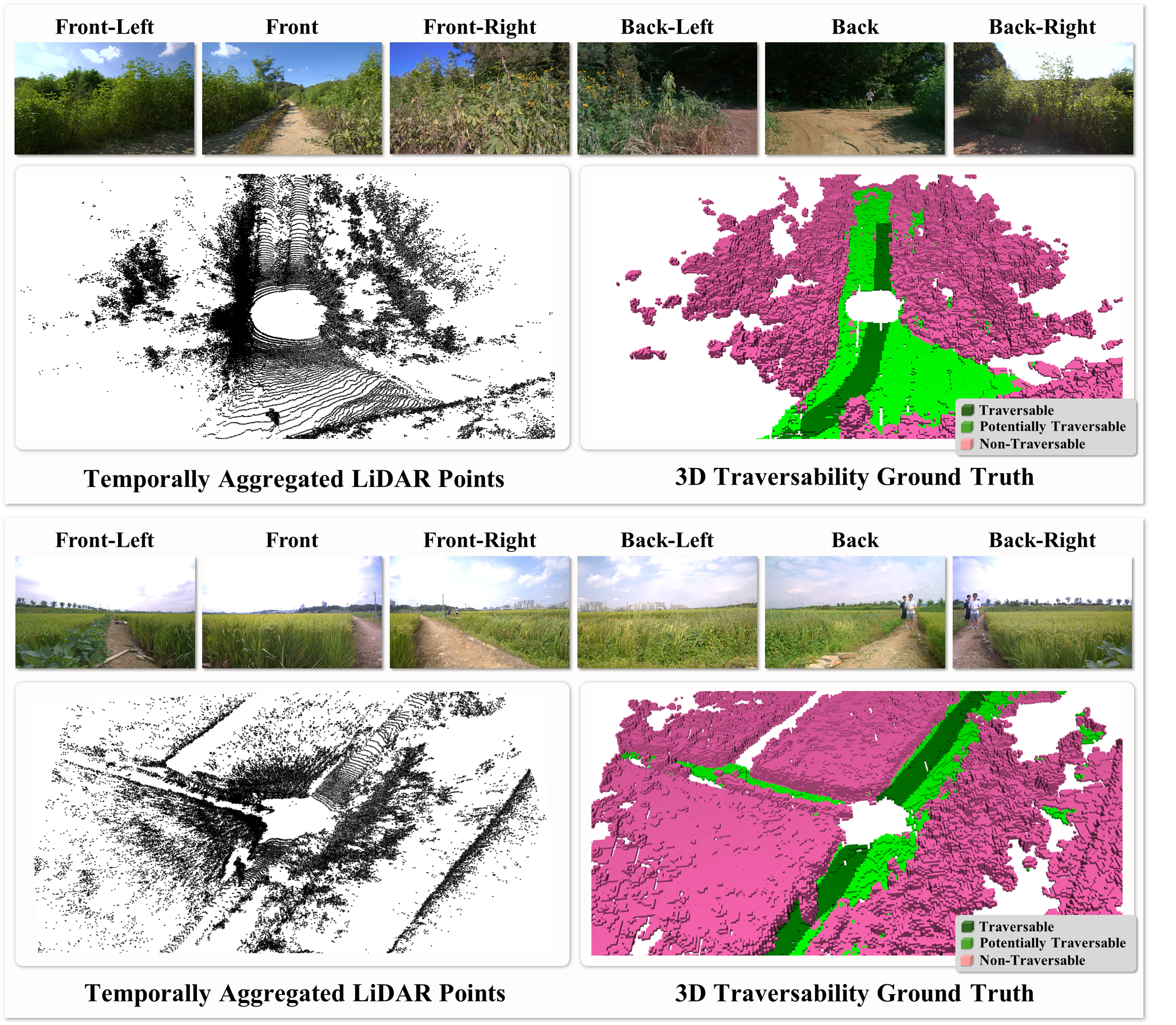}
        \label{fig:offroad_terrain}
    \caption{\textbf{Visualization of the STONE dataset.} The top shows 360-degree images of the environment captured around the robot, the bottom left presents temporally aggregated LiDAR points, and the bottom right illustrates the automatically generated 3D traversability ground-truth map.}
    \label{fig:gt_vis}
\end{figure*}

\subsection{Traversability-Oriented Feature Extraction}
For each vertex $v_i \in \mathcal{V}$ of the reconstructed mesh, we compute geometric parameters—elevation, slope, and roughness—to provide complementary cues for identifying drivable regions.

\begin{itemize}
    \item \textbf{Elevation ($h_i$)}:  A UGV may fail to traverse regions where the elevation exceeds its climbing capability. We define $h_i$ as the $z$-coordinate of vertex $v_i = (x_i, y_i, z_i)^\top$:  
    
    \begin{equation}
    \qquad h_i = z_i .
    \end{equation}

    \item \textbf{Slope ($\theta_i$)}: A UGV struggles to traverse regions with excessively steep slopes due to vehicle dynamics, such as mass, speed, and wheel ground interaction. We extract the per-vertex normal vectors $\mathbf{n}_i$ from the mesh reconstructed via Poisson surface reconstruction, where each $\mathbf{n}_i$ is computed as the normalized average of the normals of faces incident to vertex $v_i$.
    The Slope $\theta_i$ is then defined as the angle between the normal vector $\mathbf{n}_i$ and the global vertical axis $\mathbf{z} = [0,0,1]^\top$:  
    \begin{equation}
    \theta_i = \arccos\!\left( \mathbf{n}_i \cdot \mathbf{z} \right),
    \end{equation}

      
    \item \textbf{Roughness ($r_i$)}: Roughness measures local deviations from planarity on the terrain surface. 
For each vertex $v_i$, it is defined as the logarithm of the mean squared error (MSE) 
between the neighboring vertices and their best-fit plane $\Pi_i$:
\begin{equation}
    r_i = \log \!\left( \frac{1}{|N_i|} \sum_{v_j \in N_i} d(v_j, \Pi_i)^2 \right),
\end{equation}
where $N_i$ denotes the $k$-nearest neighbor set of $v_i$, $\Pi_i$ is the best-fit plane estimated 
from $N_i$, and $d(\cdot)$ represents the orthogonal distance from a vertex to the plane.
\end{itemize}

For each vertex $v_i$, we generate the geometric feature vector $\mathbf{f}_i$ as
\begin{equation}
\mathbf{f}_i = [h_i, \, \theta_i, \, r_i]^\top,
\end{equation}
The entire 3D space is discretized into a set of voxels $\mathcal{X} = \{ X_k \}_{k=1}^K$, where $K$ denotes the total number of voxels.
The geometric feature vector $\mathbf{F}_k$ for the $k$th voxel is defined as
\begin{equation}
\mathbf{F}_k = \frac{1}{|X_k|} \sum_{v_i \in X_k} \mathbf{f}_i,
\end{equation}
where $|X_k|$ is the number of vertices contained in the $k$th voxel $X_k$.

\subsection{Trajectory-Guided Traversability Auto-Labeling}

In practice, traversability is directly observed only along the robot’s trajectories, as these regions are physically traversed. However, the goal of traversability prediction is to infer traversability beyond the observed trajectories and generalize to previously unseen areas. To this end, we model the distribution of geometric features extracted from trajectory voxels as a reference distribution. Formally, the set of features along the robot’s trajectory, $\{\mathbf{F}_i\}_{i \in \mathcal{T}}$, is obtained from an underlying traversable distribution. We approximate this distribution with a multivariate Gaussian $\mathcal{N}(\boldsymbol{\mu}, \boldsymbol{\Sigma})$, where $\boldsymbol{\mu}$ and $\boldsymbol{\Sigma}$ are the empirical mean and covariance of the geometric features along the trajectory.

We define the squared Mahalanobis distance \cite{mahala} between a candidate voxel $X_k$ and the reference distribution as
\begin{equation}
D^2(X_k) = (\mathbf{F}_k-\boldsymbol{\mu})^{\top}\boldsymbol{\Sigma}^{-1}(\mathbf{F}_k-\boldsymbol{\mu}).
\end{equation}
Under the Gaussian assumption, $D^2(X_k)$ follows a chi-squared distribution with $d$ degrees of freedom, i.e., $D^2(X_k)\sim\chi^2_d$, where $d=3$ in our setting. Using the trajectory distribution as reference, we set the threshold $\chi^2_{d,1-\alpha}$, which defines the $100(1-\alpha)\%$ confidence region of traversable geometry. Voxels inside this region are considered potentially traversable, while those outside are classified as non-traversable.
\begin{itemize}
\item \textbf{Traversable ($T$)}: Voxels along the logged trajectory.
\item \textbf{Potentially Traversable ($P$)}: Off-trajectory voxels within the confidence region $\chi^2_{d,1-\alpha}$ of the trajectory distribution (e.g., $\alpha=0.05$).
\item \textbf{Non-Traversable ($N$)}: Off-trajectory voxels outside this region.
\end{itemize}

\begin{figure}[!t]
    \centering
        \centering
        \includegraphics[width=0.49\textwidth]{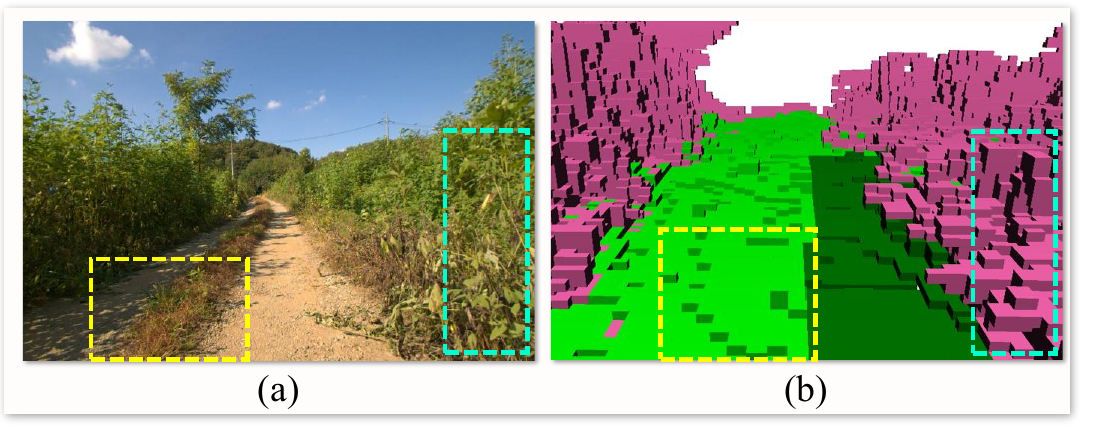}
        \label{fig:offroad_terrain}
    \caption{\textbf{Examples of regions with the same semantic class (vegetation) but different traversability labels.} (a) presents a front-view image, and (b) presents the corresponding ground-truth. Green and magenta regions denote traversable and non-traversable regions, respectively. Although the cyan and yellow boxes both correspond to vegetation, the cyan box is correctly marked as non-traversable, and the yellow box remains labeled as traversable.}
    \label{fig:gt_vis2}
\end{figure}

\subsection{Visualization of 3D Traversability}
As illustrated in Fig.~\ref{fig:gt_vis}, the generated ground-truth labels are visually coherent, clearly delineating traversable and non-traversable regions. The consistency across different scenes indicates that the automatic generation pipeline produces reliable annotations that align well with the underlying terrain geometry.

To illustrate the importance of incorporating geometric features, Fig.~\ref{fig:gt_vis2} presents a scenario in which semantic appearance alone does not reliably indicate traversability. Although both the yellow and cyan regions correspond to vegetation, their physical properties differ substantially: the low vegetation highlighted in yellow remains physically passable, whereas the dense bush in cyan obstructs any feasible path. Our traversability map (right) captures this distinction by labeling the ground vegetation as traversable (green) and the roadside bush as non-traversable (magenta).

    
\section{Benchmarks and Evaluation}
In this section, we introduce the benchmarks designed for evaluating 3D traversability prediction methods. We also present the performance results of several baseline models on these benchmarks.

\begin{table}[!t]
\caption{Evaluation of 3D traversability baselines based on IoU metrics (\%)}
\label{tab:baseline}
\setlength{\tabcolsep}{2pt}
\centering
\renewcommand{\arraystretch}{1.1}
\begin{tabular}{l|c|c|ccc|c}
\toprule[1.5pt]
Model & Sensor & IoU$_{occ}$ & IoU$_{T}$ & IoU$_{P}$ & IoU$_{N}$ & mIoU \\
\midrule
C-OpenOcc \cite{openocc}    & C   & 35.0 & 14.5 & 23.8 & 33.7 & 24.0 \\
C-CoNet \cite{openocc}      & C   & 37.0 & 14.8 & 24.3 & 34.2 & 24.4 \\
OccFormer \cite{occformer}  & C   & 37.8 & 16.3 & 29.0 & 34.1 & 26.5  \\
TPVFormer \cite{tpvformer}  & C   & 39.6 & 16.9 & 29.6 & 35.5 & 27.3 \\
\midrule
L-OpenOcc \cite{openocc}    & L   & 61.2 & 21.1 & 33.4 & 55.2 & 36.5 \\
L-CoNet \cite{openocc}      & L   & 61.6 & 22.5 & 34.8 & 57.0 & 38.1 \\
\midrule
OccFusion \cite{occfusion}  & C+R & 45.2 & 21.5 & 36.4 & 41.3 & 33.1 \\
\midrule
M-OpenOcc \cite{openocc}        & C+L & 62.3 & 18.1 & 35.0 & 61.0 & 38.0 \\
M-CoNet \cite{openocc}      & C+L & 62.6 & 18.3 & 36.0 & 62.0 & 38.8 \\
OccFusion \cite{occfusion}  & C+L & \textbf{66.1} & \textbf{19.0} & \textbf{36.9} & \textbf{62.8} & \textbf{39.6} \\
\bottomrule[1.5pt]
\multicolumn{7}{l}{R, C, and L denote Radar, Camera, and LiDAR, respectively.}
\end{tabular}%
\end{table}

\subsection{Evaluation Metrics}
The performance of 3D traversability prediction is evaluated using standard IoU-based metrics. 

\begin{itemize}
    \item \textbf{Occupancy IoU} measures the structural accuracy of free and occupied space at the voxel level:
    \begin{equation}
    \text{IoU}_{\text{occ}} = \frac{TP_{\text{occ}}}{TP_{\text{occ}} + FP_{\text{occ}} + FN_{\text{occ}}},
    \end{equation}
    where $TP_{\text{occ}}$, $FP_{\text{occ}}$, and $FN_{\text{occ}}$ denote voxel-level true positive, false positive and false negative,  respectively.

    \item \textbf{Per-class IoU} evaluates the ability to distinguish traversability classes within occupied voxels:
    \begin{equation}
    \text{IoU}_c = \frac{TP_c}{TP_c + FP_c + FN_c}, \quad c \in \{T, P, N\},
    \end{equation}
    where $TP_c$, $FP_c$, and $FN_c$ denote per-class statistics for Traversable ($T$), Potentially Traversable ($P$), and Non-Traversable ($N$).

    \item \textbf{Mean IoU (mIoU)} is the average of per-class IoUs:
    \begin{equation}
    \text{mIoU} = \frac{1}{|C|} \sum_{c\in C} \text{IoU}_c,
    \end{equation}
    where $C$ denotes the set of classes.
\end{itemize}

\subsection{Experimental Settings}
All reference methods were trained and evaluated on the splits of the STONE dataset. The traversability ground truth (GT) covers a range of [-25.6m, -25.6m, -2m, 25.6m, 25.6m, 4.4m] with a voxel size of [0.2m, 0.2m, 0.2m] in the ego coordinate system. All experiments were conducted using the PyTorch-based mmdetection3d framework on four NVIDIA RTX 3090 GPUs, and the hyperparameters of each model (e.g., optimizer, learning rate, batch size, etc.) followed the settings specified in their official implementations.

\subsection{Experimental Results}
Table~\ref{tab:baseline} presents the performance of several 3D traversability prediction methods evaluated on our STONE dataset. To provide a comprehensive benchmark, we report results across different sensing modalities, including camera-only, LiDAR-only, and multi-modal fusion (camera+LiDAR, camera+radar). The results demonstrate that incorporating complementary sensors generally improves performance over single-modality baselines, establishing a solid reference point for future research on multi-modal traversability prediction in off-road environments.

\section{CONCLUSIONS}
In this work, we introduce STONE, a scalable 3D traversability dataset designed to advance off-road autonomous navigation. While reliable off-road navigation requires a 360° field of view and robustness under adverse conditions, existing datasets fail to provide these essential capabilities. STONE integrates a 360° LiDAR, surround-view cameras, and surround-view 4D radars with precise time synchronization, and it covers diverse scenarios across varying terrains and illuminations.
Manual annotation of fine-grained semantic classes is inherently unscalable and limits the potential of data-driven perception models. To address this, we propose a fully automated pipeline, which generates 3D traversability map GTs in volumetric space. The pipeline leverages three geometric features extracted from LiDAR point clouds and uses the UGV’s traversed trajectory as supervision, thereby producing consistent, geometry-aware labels without human annotation.
We believe that STONE, together with the provided baselines, establishes a strong foundation for future research in 3D traversability prediction and will accelerate progress toward fully autonomous off-road robots.

\section{LIMITATIONS AND FUTURE WORK}
Due to budget and time constraints, the current version of STONE does not cover a wide range of off-road regions or diverse weather conditions. In future work, we plan to expand both the size and geographic scope of the dataset, as well as include data collected under challenging conditions such as snow, rain, and fog. These extensions will further enable research on robust perception and planning in adverse environments.

\section{Acknowledgement}
This work was partly supported by Institute of Information \& communications Technology Planning \& Evaluation (IITP) grant funded by the Korea government(MSIT) [NO.RS-2021-II211343, Artificial Intelligence Graduate School Program (Seoul National University)] and the National Research Foundation (NRF) funded by the Korean government (MSIT) (No. RS-2024-00421129).










\end{document}